\documentclass{article}

\usepackage[preprint]{neurips_2025}

\usepackage[utf8]{inputenc} 
\usepackage[T1]{fontenc}    
\usepackage{hyperref}       
\usepackage{url}            
\usepackage{booktabs}       
\usepackage{amsfonts}       
\usepackage{nicefrac}       
\usepackage{microtype}      
\usepackage{xcolor}         
\usepackage{amsmath}
\usepackage{subcaption}

\usepackage[capitalize]{cleveref}
\usepackage{graphicx}

\title{Latency and Token-Aware Test-Time Compute}

\author{%
Jenny Y.~Huang$^{1,3}$ \; 
Mehul Damani$^{1,3}$ \; 
Yousef El-Kurdi$^{2}$ \; 
Ramon Astudillo$^{2,3}$ \; 
Wei Sun$^{2,3}$ \\
\\
$^{1}$Department of Electrical Engineering and Computer Science, 
MIT \\
\texttt{\{jhuang9, mehul42\}@mit.edu} \\
\\
$^{2}$IBM Research, $^{3}$MIT-IBM Watson AI Lab \\
\texttt{\{yousefelk, ramon.astudillo, sunw\}@us.ibm.com} \\
}

\begin{document}

\maketitle

\begin{abstract}
Inference-time scaling has emerged as a powerful way to improve large language model (LLM) performance by generating multiple candidate responses and selecting among them. However, existing work on dynamic allocation for test-time compute typically considers only parallel generation methods such as best-of-$N$, overlooking incremental decoding methods like beam search, and has largely ignored latency, focusing only on token usage. We formulate inference-time scaling as a problem of dynamic compute allocation and method selection, where the system must decide \emph{which strategy to apply} and \emph{how much compute to allocate} on a per-query basis. Our framework explicitly incorporates both token cost and wall-clock latency, the latter being critical for user experience and particularly for agentic workflows where models must issue multiple queries efficiently. Experiments on reasoning benchmarks show that our approach consistently outperforms static strategies, achieving favorable accuracy–cost trade-offs while remaining practical for deployment.
\end{abstract}

\section{Introduction}
\label{sec:introduction}
An emerging paradigm in large language models (LLMs) is \emph{inference-time scaling}, where performance improves by allocating more computation at inference rather than through additional training. Instead of producing a single response, models generate multiple candidates using strategies such as best-of-$N$ sampling, self-consistency (majority voting), or beam search \citep{ouyang2022training,wang2023self}. This approach has proven particularly effective in reasoning-intensive domains such as mathematics and coding, where additional compute enables longer chains of thought and substantially stronger performance \citep{openai2024o1,brown2024large,guo2025deepseek}. 

Yet these gains come at steep computational cost. Fixed strategies risk overspending on simple cases while under-provisioning hard ones, a problem amplified in reasoning models  where each trace may span thousands of tokens \citep{wu2025more}. In agentic workflows with repeated queries, the burden compounds further. Prior work shows that different inference-scaling strategies perform differently depending on problem difficulty \citep{snell2024scaling}, underscoring the need not only to allocate the right amount of compute but also to select the most effective strategy. This raises a central question: \emph{can we preserve the benefits of compute-intensive inference without paying their full price?}

Several recent works aim to make inference-time scaling more efficient by adapting to query difficulty. \citet{snell2024scaling} demonstrate that adapting inference strategies by difficulty can outperform best-of-$N$ with $2$ to $4$ times less compute, but their notion of difficulty relies on oracle correctness or costly Monte Carlo sampling, making it computationally prohibitive to use in deployment. \citet{damani2024learning} offer a more practical alternative by training lightweight predictors for adaptive routing, though their method focuses on best-of-$N$ and does not test out other decoding methods.

More broadly, optimal compute allocation parallels the problem of LLM routing \citep{ong2024routellm,hu2024routerbench,tsiourvas2025causal}, where the goal is to select the most appropriate model given query heterogeneity. \citet{ding2025best} extend this idea to routing across both models and inference strategies, i.e., choosing between a large model or several smaller ones augmented with test-time compute, but their analysis is restricted to best-of-$N$.

Most work on optimal compute allocation has focused solely deciding the number of generations to sample \citep{damani2024learning,zhang2024scaling,ding2025best,raman2025abon}, despite evidence that different inference strategies behave differently across difficulty levels \citep{snell2024scaling}, e.g., best-of-$N$ excels on easy queries, beam search on harder queries. 
Crucially, however, \emph{not all tokens are generated equally}: best-of-$N$ and majority voting can exploit parallel generation of independent samples, whereas beam search requires step-by-step synchronization across beams, incurring higher latency. In interactive settings, latency is as critical as accuracy and compute budget, an aspect overlooked in prior work and a central focus of this work. 

In this work, we propose a latency- and token-aware framework for inference-time scaling that jointly determines \emph{which strategy to apply} and \emph{how much compute to allocate per query}.

\section{Methodology}
\label{sec:preliminaries}
Let $\mathcal{M}$ denote the set of inference-time scaling methods under consideration.  
Each method $m \in \mathcal{M}$ is parameterized by a vector of hyperparameters $\theta_m$.
We define a \emph{decoding strategy} as a tuple $s := (m, \theta_m)$, where $m \in \mathcal{M}$ and $\theta_m$ are method-specific hyperparameters. The system must select, on a per-query basis, the optimal strategy $s$.  While we focus on a representative subset of inference-time scaling methods that have also been studied by \citet{snell2024scaling}, the framework we propose extends to more advanced approaches \citep{yao2023tree,shinn2023reflexion,astudillo2025optimal}.  

\subsection{Inference Scaling Methods}
Sampling-based methods such as majority voting and best-of-$N$ admit \emph{parallel generation}: $N$ candidate responses can be produced simultaneously, so latency grows only modestly with $N$. Their hyperparameter is simply $\theta_m = N$, the number of candidates.  

\textbf{Majority Voting.}  Output the most frequent final answer among $N$ candidates.  

\textbf{Best-of-$N$.} Output the candidate with the highest reward score, either using raw scores (\emph{Naive}) or aggregated scores across identical responses (\emph{Weighted}).  

\textbf{Beam Search.} Unlike sampling-based methods, beam search is inherently \emph{incremental}: partial solutions are expanded step by step. At each step, up to $W$ continuations are generated for each active beam, scored by a process reward model (PRM), and the top-$N$ beams are retained. After at most $D$ steps, this yields $N$ complete solutions, from which the final answer is chosen via majority voting. The method is parameterized by $\theta_{\text{Beam}} = (N, W, D)$, where $N$ is the number of active beams, $W$ the branching factor, and $D$ the maximum depth.

\subsection{Utility Formulation}
Each decoding strategy $s$ applied to a query $x$ is characterized by three quantities:  

\textbf{Accuracy ($a_s(x)$).} A measure of output quality. Its definition depends on the domain: in mathematics, it corresponds to exact match of the predicted final answer with the ground truth; in coding, it may be the fraction of unit tests passed; and in open-ended tasks such as dialogue or summarization, it can be derived from reward model scores or human preference judgments.  

\textbf{Token Cost ($T_s(x)$).} The number of output tokens generated, reflecting the computational load.  

\textbf{Latency ($L_s(x)$).} The total wall-clock time required for decoding and reward evaluation, capturing responsiveness from the user’s perspective.

We define the utility of strategy $s$ on query $x$ as  
\begin{equation}
U_s(x) \;=\; a_s(x) \;-\; \lambda_T T_s(x) \;-\; \lambda_L L_s(x),
\label{eq:utility}
\end{equation}
where $\lambda_T, \lambda_L \ge 0$ are penalty weights, set according to user preferences, reflecting the relative importance of token usage and latency. This formulation balances two complementary dimensions of efficiency: \emph{computational load} (measured in tokens) and \emph{responsiveness} (measured in time).  

\subsection{Optimal Strategy Selection}
For each query $x$, the optimal decoding strategy is  
\[
s^*(x) \;=\; \arg\max_{s \in \mathcal{S}} U_s(x).
\]
Thus, inference-time scaling is cast as a problem of \emph{dynamic compute allocation and method selection}: determining both which strategy to apply and how much compute to allocate on a per-query basis.

\subsection{Modeling Utility}
At inference time we cannot directly observe accuracy $a_s(x)$, token count $T_s(x)$, or latency $L_s(x)$. Accuracy requires the ground truth, while token count and latency are only known after generation. To make the utility in \cref{eq:utility} computable in advance, we train predictors for these quantities.  

\textbf{Accuracy Model ($\hat{a}_s(x)$).}  
We train a lightweight probe to estimate, for query $x$ and strategy $s$, the probability of producing a correct answer, thus capturing query difficulty. Features include (i) semantic embeddings of the query and (ii) contextual descriptors of the decoding strategy (e.g., number of generations, beam width, query length). A two-layer MLP is trained with cross-entropy loss. This probe is simple to train and deploy yet yields accurate estimates of strategy accuracy (details in \cref{sec:difficulty-probe}).  

\textbf{Cost Model ($\hat{T}_s(x), \hat{L}_s(x)$).}  
For each decoding strategy, we precompute average token count and execution time from the training set. At inference, these mean values are used as predicted costs in the utility function. As shown in \cref{fig:predicted-vs-true-cost-tc,fig:predicted-vs-true-cost-late}, this approximation closely matches oracle costs, indicating that cost variation is dominated by the choice of strategy rather than the query.

\section{Experiments}
\label{sec:results}
\textbf{Setup.} We test our query-adaptive decoding strategy on the NuminaMath-CoT dataset \citep{li2024numinamath}, a large-scale benchmark of mathematical reasoning queries designed to assess step-by-step reasoning in LLMs. As the generator, we adopt Alibaba’s Qwen2.5-1.5B-Instruct, a compact instruction-tuned model specialized for mathematics. To evaluate generated solutions, we use Qwen/Qwen2.5-Math-PRM-7B, a larger process reward model (PRM) trained for mathematical reasoning supervision.

\begin{figure}[ht]
    \centering
    \begin{subfigure}[b]{0.48\textwidth}
        \centering
        \includegraphics[width=\linewidth]{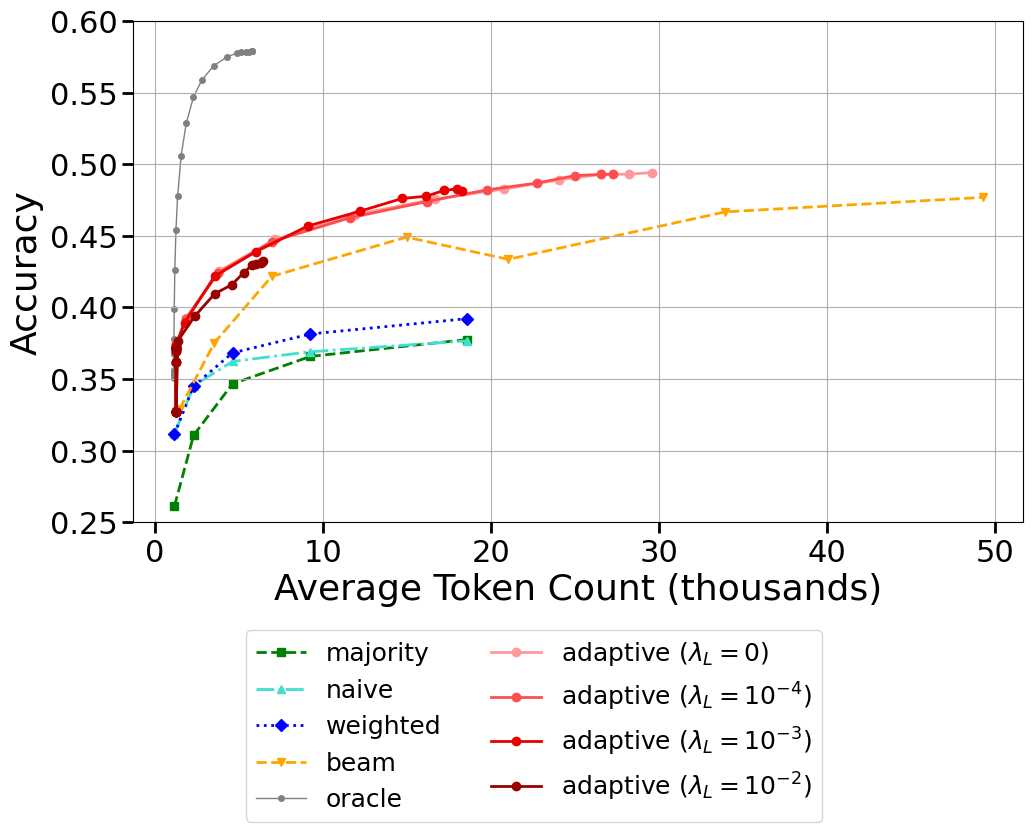}
         \caption{Latency penalty $\lambda_L$ is fixed at discrete values, while $\lambda_T$ is varied finely to capture token usage.}
        \label{fig:token-aware-router}
    \end{subfigure}
    \hspace{0.02\textwidth} 
    \begin{subfigure}[b]{0.48\textwidth}
        \centering
        \includegraphics[width=\linewidth]{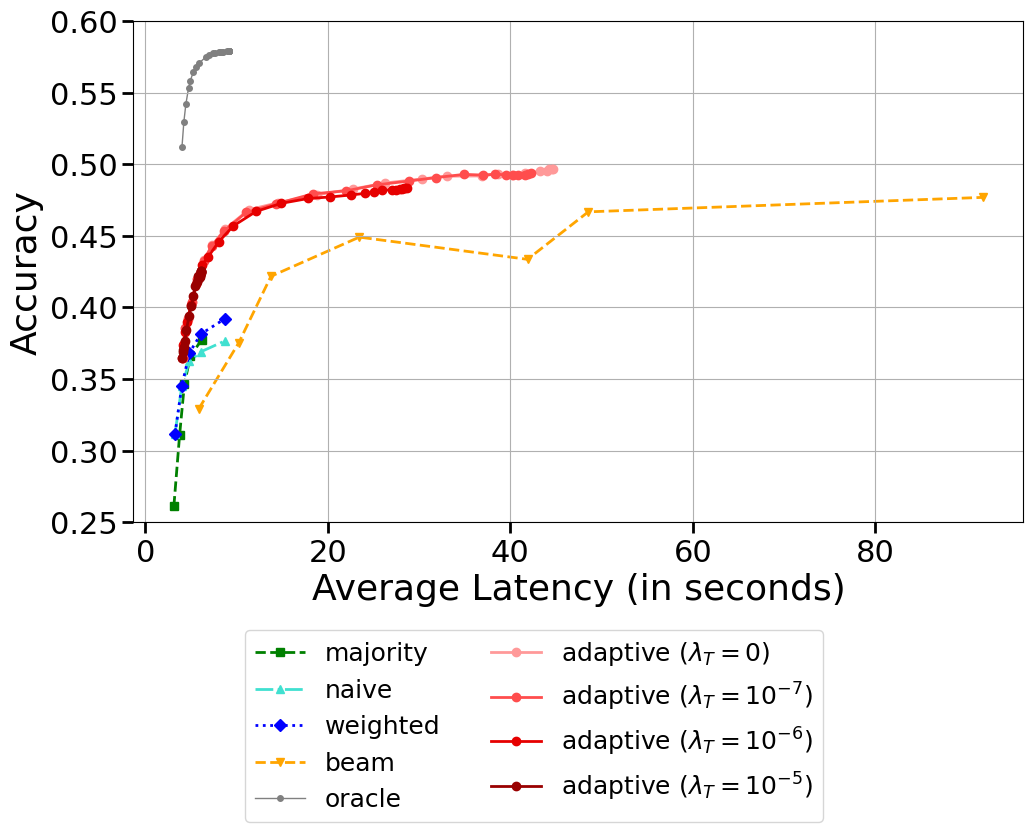}
        \caption{Token penalty $\lambda_T$ is fixed at discrete values, while $\lambda_L$ is varied more finely to capture latency effect.}
        \label{fig:latency-aware-router}
    \end{subfigure}
    \caption{Accuracy-cost trade-offs comparing the adaptive strategy to static inference-scaling methods. Accuracy is measured by soft-label correctness and cost measured as average  (a) tokens generated per query, or (b) latency. }
    \label{fig:numina-router-results}
\end{figure}
\begin{figure}[ht]
    \centering
    \includegraphics[width=0.80\textwidth]{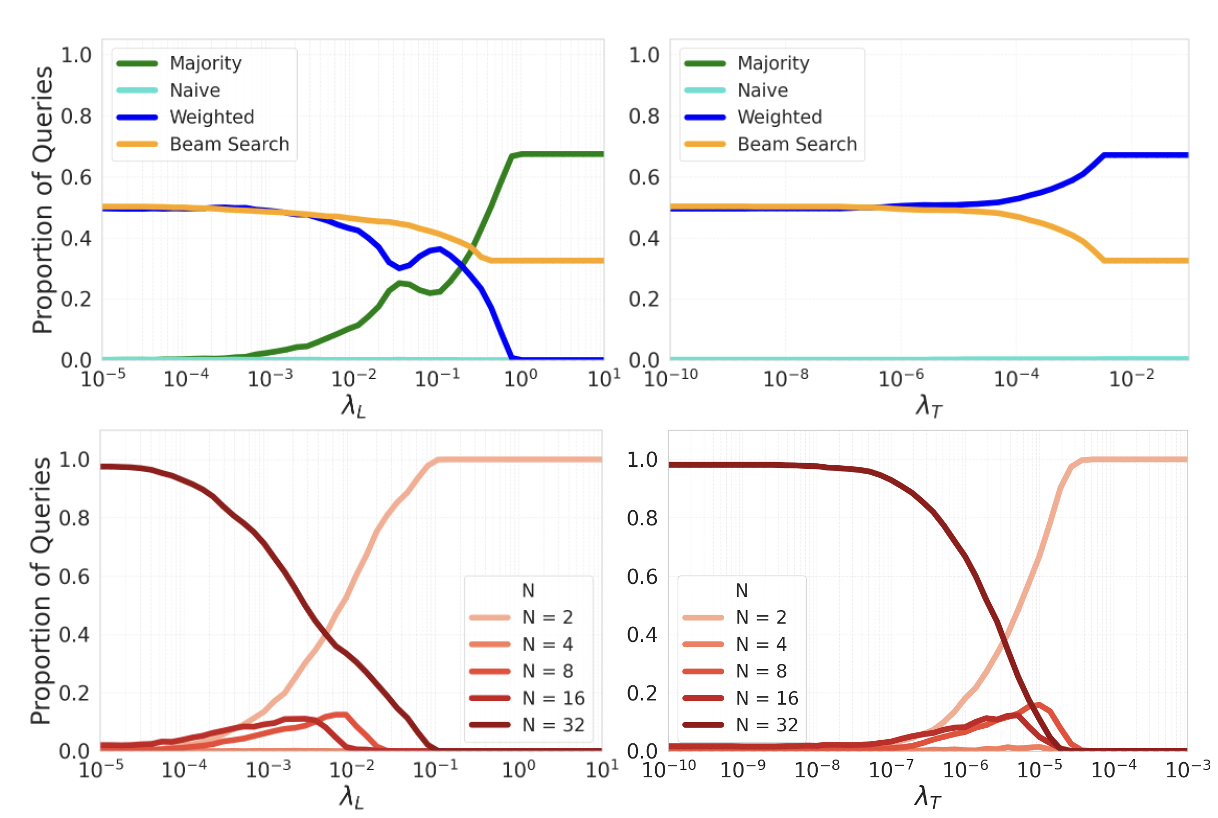}
    \caption{Selections made by the query-adaptive strategy. The top row shows the proportion of queries routed to each inference-time scaling method as the cost penalty on latency (left) and token count (right) increases. The bottom row shows the proportion of queries routed to each value of number of generations ($N$).}
    \label{fig:routing-choices}
\end{figure}

\textbf{How does the query-adaptive strategy perform?}  
From \cref{fig:numina-router-results}, we observe that the query-adaptive strategy consistently achieves superior accuracy–cost trade-offs compared to static strategies, which plateau at relatively low accuracy.

An interesting pattern emerges when both token and latency penalties are varied. For example, in \cref{fig:latency-aware-router}, with no token penalty ($\lambda_T = 0$) the adaptive strategy tends to select compute-heavy methods, reaching nearly 0.50 accuracy but at the expense of longer runtimes ($\sim$40s). Increasing $\lambda_T$ discourages token-intensive methods and shifts selection toward lighter ones, substantially reducing latency while maintaining competitive accuracy. Similar pattern  is also shown in \cref{fig:token-aware-router}.



\textbf{Which inference scaling method is being chosen?} 
Figure~\ref{fig:routing-choices} illustrates how strategy selection shifts as penalties on latency (left) and token usage (right) increase. When penalties are low, the adaptive method frequently invokes beam search, tolerating higher token counts (Figure~\ref{fig:routing-choices}, left) and longer runtimes (Figure~\ref{fig:routing-choices}, right) in exchange for accuracy. As $\lambda$ grows, reliance on expensive strategies decreases, and the method routes most queries to lighter, low-cost options. A similar trend is observed in the number of generations (Figure~\ref{fig:routing-choices}, bottom row): at low penalties, the method often selects large $N$, but it progressively shifts toward smaller $N$ as costs are weighted more heavily.

\section{Discussion}
\label{sec:discussion}

We studied inference-time scaling as a problem of dynamic compute allocation and method selection.  
Our framework jointly considers \emph{which method to apply} and \emph{how much compute to allocate}, explicitly accounting for both token cost and latency.  
The latter is especially important for user experience and represents a key extension beyond prior work that focused solely on token usage.  
Experiments on reasoning benchmarks show that query-adaptive selection achieves substantially better accuracy–efficiency trade-offs than fixed strategies.  
Beyond the methods explored here, our formulation extends naturally to more advanced inference-time scaling techniques, and future work will evaluate broader domains such as coding and dialogue.  
Another promising direction is to close the remaining gap between the adaptive strategy and the oracle by improving the accuracy of the probe used to estimate utility.  
We believe this line of work will be particularly impactful for agentic workflows, where models must issue multiple queries and efficiency becomes critical.


\newpage
\bibliography{iclr2025_conference}
\bibliographystyle{iclr2025_conference}

\appendix
\section{Appendix}
\label{sec:appendix}

\label{sec:difficulty-probe}
\subsection{Accuracy Probe.} 

\paragraph{Data Collection.}
To construct training data for the accuracy probe, we run multiple decoding strategies on a subset of NuminaMath CoT and record the results. For each input query, $x$, we execute a range of decoding strategies. Each run produces a set of candidate completions, which we label as correct or incorrect by comparing the extracted answer against the ground truth. LLM generation is stochastic, so a single run does not provide a stable estimate of method performance on a query. To obtain a more reliable supervision signal, we repeatedly sample completions from each decoding strategy to compute the empirical accuracy as the fraction of outputs matching the ground truth. This fraction serves as a \emph{soft label}, providing a continuous value between 0 and 1 that reflects the expected success probability of strategy $s$ on query $x$.

The resulting dataset consists of tuples,
\[
(s, x, \text{features}(s, x), \hat{a}_s(x))
\]
where $\text{features}(s, x)$ include both query-level embeddings and strategy parameters, and $\hat{a}_s(x)$ is the soft accuracy label obtained via repeated sampling.

\paragraph{Embedding Backbones.}
We experiment with two different types of embeddings for encoding input queries:

\texttt{Qwen2.5-1.5B-Instruct.} We pass each input through the generator model once and apply max pooling across the final hidden states to obtain a 1536-dimensional vector.

\texttt{BERT-base-uncased.} We extract the 768-dimensional hidden state of the [CLS] token, following the conventional practice for sentence-level tasks.

\paragraph{Contextual Features.}
In addition to embeddings, we concatenate the following features:
\begin{itemize}
    \item decoding parameters: number of generations, beam size, beam width, maximum iteration count.
    \item method type: encoded as one-hot vectors for beam search, best-of-N, majority voting, naive, or weighted approaches.
    \item query-level metadata: problem length (number of tokens).
\end{itemize}

\paragraph{Model Architecture.}
The probe is a two-layer MLP with hidden dimensions of 200–200–1. GELU activations are applied between layers. The output layer produces a scalar logit, which is transformed into a probability via the logistic function.

\paragraph{Training Dynamics.}
The probe is optimized using Adam with a learning rate of $10^{-5}$, and trained against soft accuracy labels using a binary cross-entropy loss with logits. We train for up to 10 epochs with early stopping based on validation loss (patience = 1).

\paragraph{Calibration.}
To ensure well-calibrated predictions, we apply Platt scaling on a held-out calibration set. We found this step to improve the alignment between predicted and empirical accuracies.
\begin{figure}[ht]
    \centering
    \includegraphics[width=0.50\textwidth]{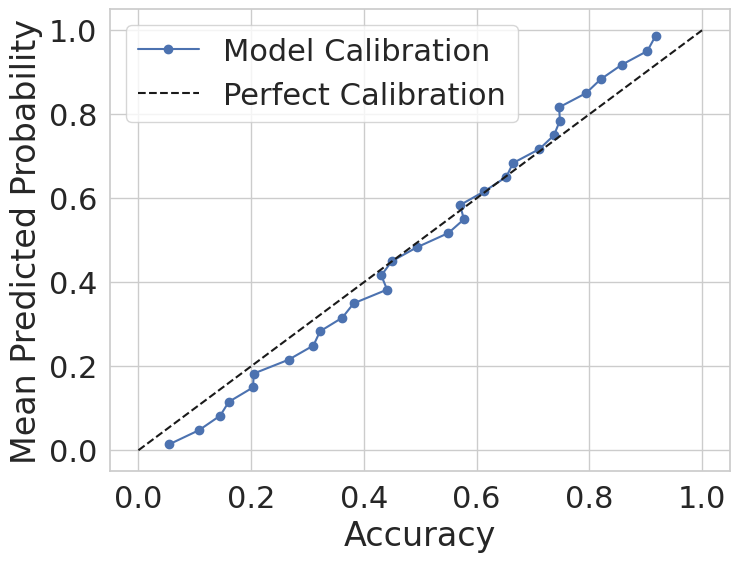}
    \caption{The accuracy model achieves strong calibration with ground-truth correctness rates.}
    \label{fig:roc-curves-calibration}
\end{figure}
Our resulting accuracy model achieves strong calibration with ground-truth correctness rates \cref{fig:roc-curves-calibration}.

\subsection{Cost Measurements.}
\label{sec:cost-measurements}

\subsubsection{Latency.}
For each query, latency is measured as the total wall-clock time required for both generation and scoring under the chosen decoding strategy.

\textbf{Generation Time.} We measure wall-clock latency for completion generation using vLLM. For each user query, we duplicate the templated prompt N times so that a single batch corresponds to one original query with N candidate completions. We set the generation batch size equal to the candidate count (i.e., batch size = $N$), which yields one vLLM generate call per query. For each query batch, we record the wall-clock time surrounding the call to llm.generate(). Scoring latency is measured starting from after the LLM is already loaded into memory.

\textbf{Scoring Time.} Scoring latency is measured as the wall-clock time taken to score a list of candidate completions step, from the moment all candidates are available until the PRM returns scores. Scoring latency is measured starting from after the PRM is already loaded into memory.

\textbf{Hardware.} All experiments are run on a single GPU NVIDIA A100 80GB.

\subsubsection{Token Count.}
We record the length of the tokenized output sequence (using the model’s tokenizer) across all sequences generated throughout the run of a decoding strategy and aggregate these to obtain the total \textit{token count}. 
\begin{figure}[h]
    \centering
    \includegraphics[width=0.95\textwidth]{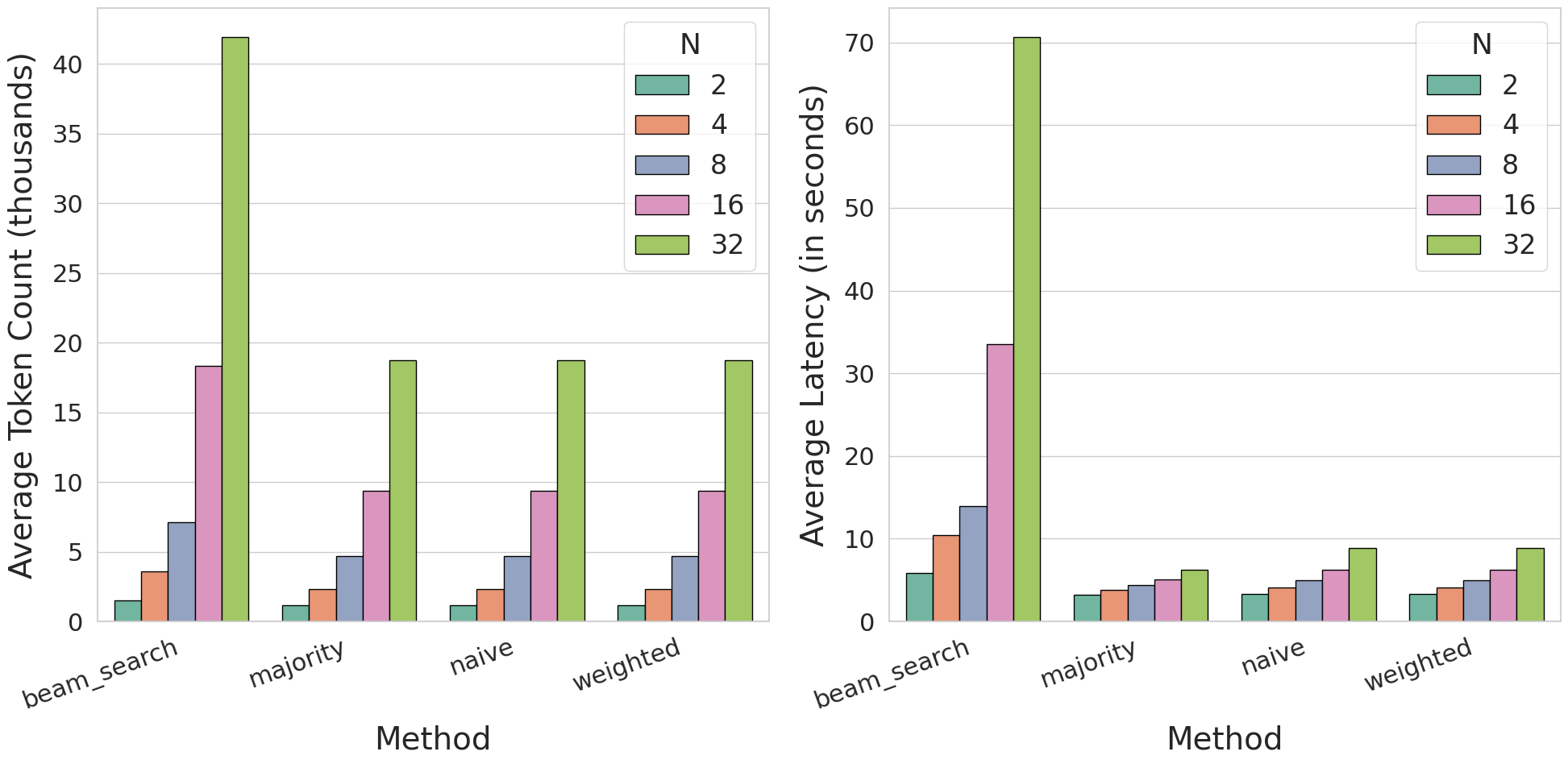}
    \caption{Different decoding procedures lead to differences in compute costs (both token-count and wall-clock latency). For instance, we find that while beam-search is the highest accuracy method, it uses drastically more compute than the other methods, which calls for the need to route depending on query difficulty.}
    \label{fig:tokens-and-latencies}
\end{figure}

\subsection{Query-Adaptive Strategy with BERT Embeddings for Accuracy Probe.}
We also experiment with using BERT embeddings for the accuracy probe.  
Compared to embeddings from \texttt{Qwen2.5-1.5B-Instruct} (1536 dimensions), BERT provides a more compact 768-dimensional representation.  
Although the query-adaptive strategy with BERT embeddings yields slightly lower performance, it still consistently outperforms static methods, demonstrating the robustness of our approach to the choice of embedding model.

\begin{figure}[ht]
    \centering
    \includegraphics[width=0.70\textwidth]{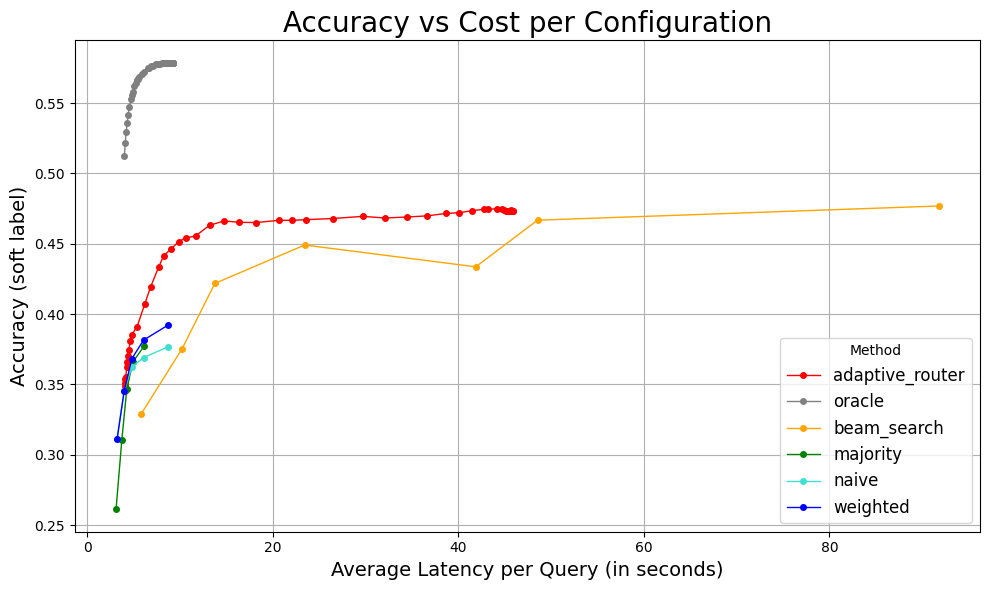}
    \caption{Accuracy–token trade-off for multiple inference-time scaling strategies when using BERT embeddings for training the difficulty probe.}
    \label{fig:adaptive-router-token-count-bert}
\end{figure}
\begin{figure}[ht]
    \centering
    \includegraphics[width=0.70\textwidth]{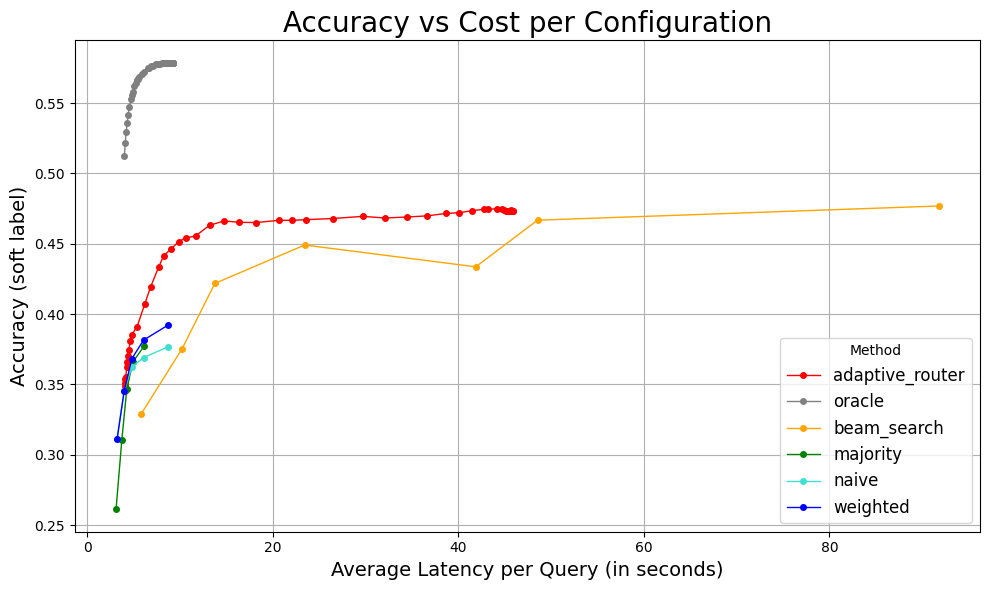}
    \caption{Accuracy–latency trade-off for multiple inference-time scaling strategies when using BERT embeddings for training the difficulty probe.}
    \label{fig:adaptive-router-latency-bert}
\end{figure}

\subsection{Query-Adaptive Strategy with Estimated vs. Ground-Truth Costs.}
\begin{figure}[ht]
    \centering
    \includegraphics[width=0.70\textwidth]{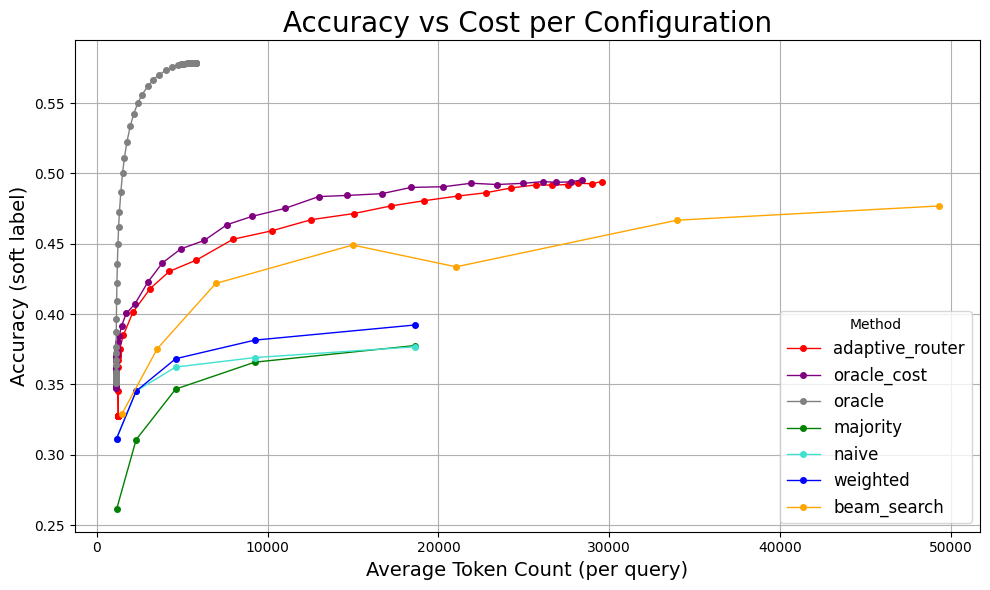}
    \caption{The adaptive strategy using the predicted token count (red) closely matches the performance achieved when using the true token count values (purple), indicating that the cost model provides reliable estimates.}
    \label{fig:predicted-vs-true-cost-tc}
\end{figure}

\begin{figure}[ht]
    \centering
    \includegraphics[width=0.70\textwidth]{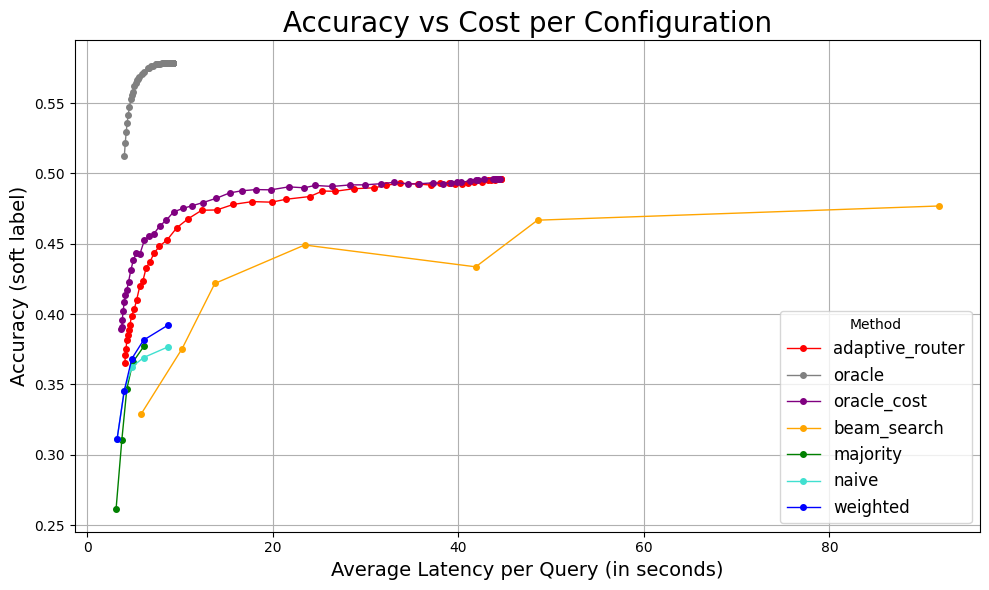}
    \caption{The adaptive strategy using the predicted latency (red) closely matches the performance achieved when using the true latency values (purple), indicating that the cost model provides reliable estimates.}
    \label{fig:predicted-vs-true-cost-late}
\end{figure}

Figure \ref{fig:predicted-vs-true-cost-tc} and \ref{fig:predicted-vs-true-cost-late} show that the adaptive strategy using predicted token counts closely tracks the performance achieved with ground-truth values, with only minor degradation, indicating that the aggregate cost model provides reliable estimates on reasoning benchmarks.


\subsection{Query-Adaptive Test Compute Strategy for Beam Search.}

\begin{figure}[ht]
    \centering
    \includegraphics[width=0.7\textwidth]{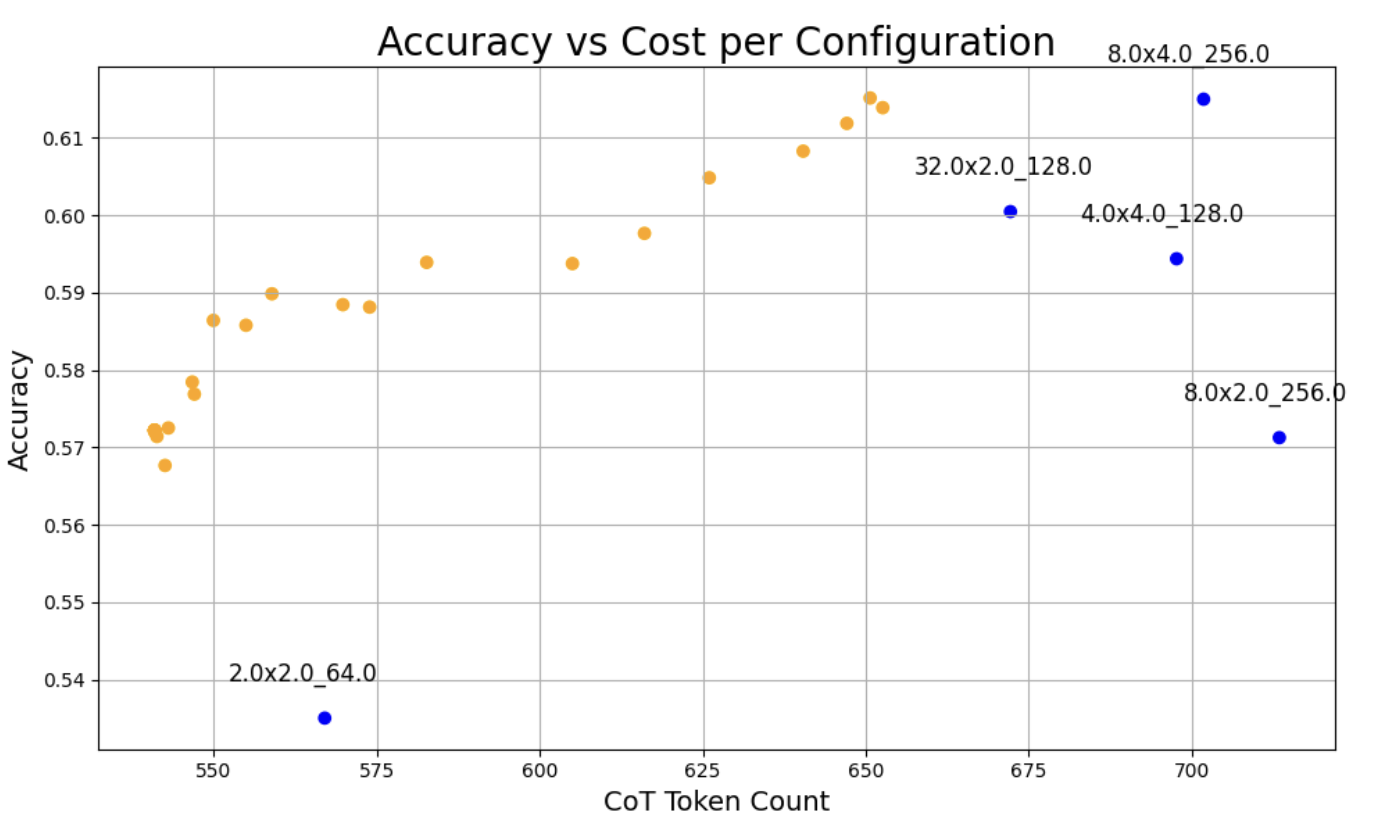}
    \caption{Accuracy-token count trade-off for multiple decoding configurations of beam search. Each point represents a decoding configuration; yellow points represent the input-adaptive strategy, while blue points represent labeled static beam search configurations, labeled with (beam size, beam width, and chunk size). The adaptive strategy is able to achieve higher accuracy at lower cost compared to the static decoding settings.}
    \label{fig:adaptive-router-beam}
\end{figure}

Instead of selecting among different inference-scaling methods, here we evaluate the adaptive strategy in a single-method setting. Focusing on beam search, the strategy selects hyperparameters (beam size, width, and chunk size) to maximize utility under a user-defined budget constraint. On Math-500 (\cref{fig:adaptive-router-beam}), the adaptive strategy achieves higher accuracy at lower cost compared to fixed decoding configurations, demonstrating the effectiveness of utility-based adaptation even within a single inference-scaling method.

\end{document}